\documentclass[11pt]{article}
\usepackage{acl2016}
\usepackage{times}
\usepackage{latexsym}
\usepackage{amsfonts}
\usepackage{color}
\usepackage{qtree}
\usepackage{url}

\usepackage{array,multirow,graphicx}

\aclfinalcopy 

\newcommand{\s}{{\cal S}}
\newcommand{\p}{{\cal P}}
\newcommand{\internal}{{\cal I}}
\newcommand{\n}{{\cal N}}

\newcommand{\ignore}[1]{}

\title{Optimizing Spectral Learning for Parsing} 

\author{Shashi Narayan and
  Shay B. Cohen \\ School of Informatics \\ University of Edinburgh \\
  Edinburgh, EH8 9LE, UK \\ \texttt{\{snaraya2,scohen\}@inf.ed.ac.uk}
}

\date{}

\begin{document}

\maketitle

\begin{abstract}
We describe a search algorithm for optimizing the number of latent
states when estimating latent-variable PCFGs with spectral methods.
Our results show that contrary to the common belief that the number of
latent states for each nonterminal in an L-PCFG can be decided in
isolation with spectral methods, parsing results significantly improve
if the number of latent states for each nonterminal is globally
optimized, while taking into account interactions between the
different nonterminals. In addition, we contribute an empirical
analysis of spectral algorithms on eight morphologically rich
languages: Basque, French, German, Hebrew, Hungarian, Korean, Polish
and Swedish. Our results show that our estimation consistently
performs better or close to coarse-to-fine expectation-maximization
techniques for these languages.
\end{abstract}

\section{Introduction}

Latent-variable probabilistic context-free grammars (L-PCFGs) have
been used in the natural language processing community (NLP) for
syntactic parsing for over a decade. They were introduced in the NLP
community by \newcite{matsuzaki-05} and \newcite{prescher2005head},
with Matsuzaki et al. using the expectation-maximization (EM)
algorithm to estimate them. Their performance on syntactic parsing of
English at that stage lagged behind state-of-the-art parsers.

\newcite{petrov-06} showed that one of the reasons that the EM
algorithm does not estimate state-of-the-art parsing models for
English is that the EM algorithm does not control well for the model
size used in the parser -- the number of latent states associated with
the various nonterminals in the grammar. As such, they introduced a
coarse-to-fine technique to estimate the grammar. It splits and merges
nonterminals (with latent state information) with the aim to optimize
the likelihood of the training data.
Together with other types of fine tuning of the parsing model, this
led to state-of-the-art results for English parsing.

In more recent work, \newcite{cohen-12} described a different family
of estimation algorithms for L-PCFGs. This so-called ``spectral''
family of learning algorithms is compelling because it offers a
rigorous theoretical analysis of statistical convergence, and
sidesteps local maxima issues that arise with the EM algorithm.

While spectral algorithms for L-PCFGs are compelling from a
theoretical perspective, they have been lagging behind in their
empirical results on the problem of parsing.
In this paper we show that one of the main reasons for that is that
spectral algorithms require a more careful tuning procedure for the
number of latent states than that which has been advocated for until
now.  In a sense, the relationship between our work and the work of
\newcite{cohen-13b} is analogous to the relationship between the work
by \newcite{petrov-06} and the work by \newcite{matsuzaki-05}: we
suggest a technique for optimizing the number of latent states for
spectral algorithms, and test it on eight languages.

Our results show that when the number of latent states is optimized
using our technique, the parsing models the spectral algorithms yield
perform significantly better than the vanilla-estimated models,
and for most of the languages -- better than the Berkeley
parser of \newcite{petrov-06}.

As such, the contributions of this parser are two-fold:

\begin{itemize}

\item We describe a search algorithm for optimizing the number of
  latent states for spectral learning.

\item We describe an analysis of spectral algorithms on eight
  languages (until now the results of L-PCFG estimation with spectral
  algorithms for parsing were known only for English). Our parsing
  algorithm is rather language-generic, and does not require
  significant linguistically-oriented adjustments.


\end{itemize}

In addition, we dispel the common wisdom that more data is needed
with spectral algorithms. Our models yield high performance on
treebanks of varying sizes from 5,000 sentences (Hebrew and Swedish)
to 40,472 sentences (German).


The rest of the paper is organized as follows.  In
\S\ref{section:background} we describe notation and
background. \S\ref{sec:optspectralalgo} further investigates the need
for an optimization of the number of latent states in spectral
learning and describes our optimization algorithm, a search algorithm
akin to beam search. In \S\ref{sec:exp} we describe our experiments
with natural language parsing for Basque, French, German, Hebrew,
Hungarian, Korean, Polish and Swedish. We conclude in
\S\ref{sec:conclusion}.

\section{Background and Notation}
\label{section:background}

We denote by $[n]$ the set of integers $\{ 1, \ldots, n \}$.
An L-PCFG is a 5-tuple $(\n, \internal, \p, f, n)$ where:

\begin{itemize}

\item $\n$ is the set of nonterminal symbols in the grammar.
  $\internal \subset \n$ is a finite set of {\em interminals}.  $\p
  \subset \n$ is a finite set of {\em preterminals}.  We assume that
  $\n = \internal \cup \p$, and $\internal \cap \p= \emptyset$.  Hence
  we have partitioned the set of nonterminals into two subsets.

\item $f \colon \n \rightarrow \mathbb{N}$ is a function that maps
  each nonterminal $a$ to the number of latent states it uses. The set
  $[m_a]$ includes the possible hidden states for nonterminal $a$.

\item $[n]$ is the set of possible words.

\item For all $a \in \internal$, $b \in \n$, $c \in \n$, $h_1 \in
  [m_a]$, $h_2 \in [m_b]$, $h_3 \in [m_c]$, we have a binary
  context-free rule $ a(h_1) \rightarrow b(h_2) \;\; c(h_3) $.

\item For all $a \in \p$, $h \in [m_a]$, $x \in [n]$, we have a
  lexical context-free rule $ a(h) \rightarrow x $.

\end{itemize}

The estimation of an L-PCFG requires an assignment of probabilities
(or weights) to each of the rules $a(h_1) \rightarrow b(h_2) \;\;
c(h_3)$ and $ a(h) \rightarrow x $, and also an assignment of starting
probabilities for each $a(h)$, where $a \in \internal$ and $h \in
[m_a]$.  Estimation is usually assumed to be done from a set of parse
trees (a treebank), where the latent states are not included in the
data -- only the ``skeletal'' trees which consist of nonterminals in
$\n$.

L-PCFGs, in their symbolic form, are related to regular tree grammars,
an old grammar formalism, but they were introduced as statistical
models for parsing with latent heads more recently by
\newcite{matsuzaki-05} and \newcite{prescher2005head}.  Earlier work
about L-PCFGs by \newcite{matsuzaki-05} used the
expectation-maximization (EM) algorithm to estimate the grammar
probabilities. Indeed, given that the latent states are not observed,
EM is a good fit for L-PCFG estimation, since it aims to do learning
from incomplete data. This work has been further extended by
\newcite{petrov-06} to use EM in a coarse-to-fine fashion: merging and
splitting nonterminals using the latent states to optimize the number
of latent states for each nonterminal.

\newcite{cohen-12} presented a so-called spectral algorithm to estimate
L-PCFGs. This algorithm uses linear-algebraic
procedures such as singular value decomposition (SVD) during learning.
The spectral algorithm of Cohen et al. builds on an estimation
algorithm for HMMs by \newcite{hsu09}.\footnote{A related algorithm
  for weighted tree automata (WTA) was developed by \newcite{bailly-10}.
  However, the conversion from L-PCFGs to WTA is not straightforward,
  and information is lost in this conversion. See also
  \cite{rabusseau-16}.}  
\newcite{cohen-13b} experimented with this spectral algorithm for parsing
English. A different variant of a spectral learning
algorithm for L-PCFGs was developed by \newcite{cohen-14b}.  It breaks
the problem of L-PCFG estimation into multiple convex optimization
problems which are solved using EM.

The family of L-PCFG spectral learning algorithms was further extended
by \newcite{narayan-15}.  They presented a simplified version of the
algorithm of \newcite{cohen-12} that estimates sparse grammars and
assigns probabilities (instead of weights) to the rules in the
grammar, and as such does not suffer from the problem of negative
probabilities that arise with the original spectral algorithm (see
discussion in Cohen et al., 2013\nocite{cohen-13b}). In this paper, we
use the algorithms by \newcite{narayan-15} and \newcite{cohen-12}, and
we compare them against state-of-the-art L-PCFG parsers such as the
Berkeley parser \cite{petrov-06}. We also compare our algorithms to
other state-of-the-art parsers where elaborate
linguistically-motivated feature specifications \cite{hall2014less},
annotations \cite{crabbe:2015:EMNLP} and formalism conversions
\cite{fernandezgonzalez-martins:2015} are used.

\section{Optimizing Spectral Estimation}
\label{sec:optspectralalgo}

In this section, we describe our optimization algorithm and its
motivation.

\subsection{Spectral Learning of L-PCFGs and Model Size}
\label{section:spectral}

\begin{figure}
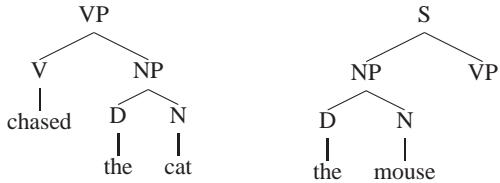

  \begin{center}
    \begin{footnotesize}
      \begin{tabular}{lp{0.3in}l}
        \Tree [.VP [.V chased ] [.NP [.D the ] [.N cat ] ] ]
        &
        &
        \Tree [.S [.NP [.D the ] [.N mouse ] ] VP ]
      \end{tabular}
    \end{footnotesize}
  \end{center}
  \caption{The inside tree (left) and outside tree (right) for the
    nonterminal {\tt VP} in the parse tree {\tt (S (NP (D the) (N
      mouse)) (VP (V chased) (NP (D the) (N cat))))} for the sentence
    {\it ``the mouse chased the cat.''}}
  \label{fig:iotrees}
\end{figure}

The family of spectral algorithms for latent-variable PCFGs rely on
feature functions that are defined for {\em inside} and {\em outside}
trees.  Given a tree, the inside tree for a node contains the entire
subtree below that node; the outside tree contains everything in the
tree excluding the inside tree. Figure~\ref{fig:iotrees} shows an
example of inside and outside trees for the nonterminal {\tt VP} in
the parse tree of the sentence {\it ``the mouse chased the cat''}.

With L-PCFGs, the model dictates that an inside tree and an outside
tree that are connected at a node are statistically conditionally
independent of each other given the node label and the latent state
that is associated with it. As such, one can identify the distribution
over the latent states for a given nonterminal $a$ by using the
cross-covariance matrix of the inside and the outside trees,
$\Omega^a$. For more information on the definition of this
cross-covariance matrix, see \newcite{cohen-12} and
\newcite{narayan-15}.

The L-PCFG spectral algorithms use singular value decomposition (SVD)
on $\Omega^a$ to reduce the dimensionality of the feature
functions. If $\Omega^a$ is computed from the true L-PCFG distribution
then the rank of $\Omega^a$ (the number of non-zero singular values)
gives the number of latent states according to the model.

In the case of estimating $\Omega^a$ from data generated from an
L-PCFG, the number of latent states for each nonterminal can be
exposed by capping it when the singular values of $\Omega^a$ are
smaller than some threshold value. This means that spectral
algorithms give a natural way for the selection of the number of latent states
for each nonterminal $a$ in the grammar.

However, when the data from which we estimate an L-PCFG model are not drawn
from an L-PCFG (the model is ``incorrect''), the number of non-zero
singular values (or the number of singular values which are large) is
no longer sufficient to determine the number of latent states for each
nonterminal.  This is where our algorithm comes into play: it
optimizes the number of latent search for each nonterminal by applying
a search algorithm akin to beam search.

\subsection{Optimizing the Number of Latent States}
\label{sec:optalgo}

\begin{figure}[th!]
  \begin{footnotesize}
    \framebox{\parbox{\columnwidth}{ {\bf Inputs:} An input treebank
        divided into training and development set.  A basic spectral
        estimation algorithm $\s$ with its default setting. An integer
        $k$ denoting the size of the beam.  An integer $m$ denoting
        the upper bound on the number of latent states.
        
        $\,$
        
        {\bf Algorithm:}
        
        (Step 0: Initialization)
        
        \begin{itemize}
          
        \item Set $Q$, a queue of size $k$, to be empty.
          
        \item Estimate an L-PCFG $G_{\s}: (\n, \internal, \p, f_{\s},
          n)$ using $\s$.
          
        \item Initialize $f=f_{\s}$, a function that maps each
          nonterminal $a \in \n$ to the number of latent states.
          
        \item Let $L$ be a list of nonterminals $( a_1,\ldots,a_M )$
          such that $a_i \in \n$ for which to optimize the number of
          latent states.

        \item Let $s$ be the $F_1$ score for the above L-PCFG $G_{\s}$
          on the development set.
          
        \item Put in $Q$ the element $(s, 1, f, \mathrm{coarse})$.
          
        \item The queue is ordered by $s$, the first element of
          tuples, in the queue.
          
        \end{itemize}
        
        (Step 1: Search, repeat until termination happens)
        
        \begin{itemize}
          
        \item Dequeue the queue into $(s, j, f, t)$ where $j$ is the
          index in the input nonterminal list $L$.
          
        \item If $j = (M+1)$, return $f$.

        \item If $t$ is $\mathrm{coarse}$ then for each $m_0 \in \{ 1,
          5, 10, \ldots, m \}$:
          
          \begin{itemize}
            
          \item Let $f_0$ be such that $\forall a \neq a_j$ $f_0(a) =
            f(a)$ and $f_0(a_j) = m_0$.
            
          \item Train an L-PCFG $G_0$ using $\s$ but with $f_0$.
            
          \item Let $s_0$ be the $F_1$ score for $G_0$ on the
            development set.
            
          \item Enqueue into $Q$:  $(s_0, j, f_0, \mathrm{refine})$.
            
          \end{itemize}
          
          
        \item If $t$ is $\mathrm{refine}$ then for each $m_0 \in \{
          f(a) + \ell \mid \ell \in \{ -4, -3, -2, -1, 0, 1, 2, 3, 4\}
          \}$:
          
          \begin{itemize}
            
          \item Let $f_0$ be such that $\forall a \neq a_j$ $f_0(a) =
            f(a)$ and $f_0(a_j) = m_0$.
            
          \item Train an L-PCFG $G_0$ using $\s$ but with $f_0$.
            
          \item Let $s_0$ be the $F_1$ score for $G_0$ on the
            development set.
            
          \item Enqueue into $Q$: $(s_0, j+1, f_0, \mathrm{coarse})$.
            
          \end{itemize}
          
        \end{itemize}
        
      }
    }
  \end{footnotesize}
  \caption{A search algorithm for finding the optimal number of latent
    states.\label{fig:algorithm}}
  \label{fig:mlelearn}\end{figure}

As mentioned in the previous section, the number of non-zero singular
values of $\Omega^a$ gives a criterion to determine the number of
latent states $m_a$ for a given nonterminal $a$.  In practice, we cap
$m_a$ not to include small singular values which are close to 0,
because of estimation errors of $\Omega^a$.

This procedure does not take into account the interactions that exist
between choices of latent state numbers for the various nonterminals.
In principle, given the independence assumptions that L-PCFGs make,
choosing the nonterminals based only on the singular values is
``statistically correct.''  However, because in practice the modeling
assumptions that we make (that natural language parse trees are drawn
from an L-PCFG) do not hold, we can improve further the accuracy of
the model by taking into account the nonterminal interaction. Another
source of difficulty in choosing the number of latent states based the
singular values of $\Omega^a$ is sampling error: in practice, we are
using data to estimate $\Omega^a$, and as such, even if the model is
correct, the rank of the estimated matrix does not have to correspond
to the rank of $\Omega^a$ according to the true distribution. As a
matter of fact, in addition to neglecting small singular values, the
spectral methods of \newcite{cohen-13b} and \newcite{narayan-15} also
cap the number of latent states for each nonterminal to an upper bound
to keep the grammar size small.

\newcite{petrov-06} improves over the estimation described in
\newcite{matsuzaki-05} by taking into account the interactions between
the nonterminals and their latent state numbers in the training
data. They use the EM algorithm to split and merge nonterminals using
the latent states, and optimize the number of latent states for each
nonterminal such that it maximizes the likelihood of a training
treebank. Their refined grammar successfully splits nonterminals to
various degrees to capture their complexity. We take the analogous
step with spectral methods. We propose an algorithm where we first
compute $\Omega^a$ on the training data and then we optimize the
number of latent states for each nonterminal by optimizing the
PARSEVAL metric \cite{black-91} on a development set.

Our optimization algorithm appears in Figure~\ref{fig:algorithm}.  The
input to the algorithm is training and development data in the form of
parse trees, a basic spectral estimation algorithm $\s$ in its default
setting, an upper bound $m$ on the number of latent states that can be
used for the different nonterminals and a beam size $k$ which gives a
maximal queue size for the beam. The algorithm aims to learn a
function $f$ that maps each nonterminal $a$ to the number of latent
states. It initializes $f$ by estimating a default grammar $G_{\s}:
(\n, \internal, \p, f_{\s}, n)$ using $\s$ and setting $f=f_{\s}$. It
then iterates over $a \in \n$, improving $f$ such that it optimizes
the PARSEVAL metric on the development set.

The state of the algorithm includes a queue that consists of tuples of
the form $(s,j,f,t)$ where $f$ is an assignment of latent state
numbers to each nonterminal in the grammar, $j$ is the index of a
nonterminal to be explored in the input nonterminal list $L$, $s$ is
the $F_1$ score on the development set for a grammar that is estimated
with $f$ and $t$ is a tag that can either be $\mathrm{coarse}$ or
$\mathrm{refine}$.

The algorithm orders these tuples by $s$ in the queue, and iteratively
dequeues elements from the queue.  Then, depending on the label $t$,
it either makes a refined search for the number of latent states for
$a_j$, or a more coarse search. As such, the algorithm can be seen as
a variant of a beam search algorithm.

\begin{table*}[htbp]
  {
    \begin{center}{\small
        \begin{tabular}{|l|l||c|c|c|c|c|c|c|c|c|}
          \hline 
          
          \multicolumn{2}{|c||}{lang.} & Basque & French & German-N & German-T & Hebrew &
          Hungarian & Korean & Polish & Swedish \\ \hline
          \parbox[t]{1mm}{\multirow{4}{*}{\rotatebox[origin=c]{90}{train}}}
          & sent. & 7,577 & 14,759 & 18,602 & 40,472 & 5,000 & 8,146 & 23,010 & 6,578 & 5,000 \\ 
          & tokens & 96,565 & 443,113 & 328,531 & 719,532 & 128,065 & 170,221 & 301,800 & 66,814 & 76,332 \\ 
          & lex. size & 25,136 & 27,470 & 48,509 & 77,219 & 15,971 & 40,775 & 85,671 & 21,793 & 14,097 \\ 
          & \#nts &  112 &  222 &  208 &  762 &  375 & 112 & 352 & 198 & 148 \\ \hline
          \parbox[t]{1mm}{\multirow{2}{*}{\rotatebox[origin=c]{90}{dev}}}
          & sent. & 948 & 1,235 & 1,000 & 5,000 & 500 & 1,051 & 2,066 &  821 & 494 \\ 
          & tokens & 13,893 & 38,820 & 17,542 & 76,704 & 11,305 & 30,010 & 25,729 &  8,391 & 9,339 \\ \hline
          \parbox[t]{1mm}{\multirow{2}{*}{\rotatebox[origin=c]{90}{test}}}
          & sent. &  946 & 2,541 & 1,000 & 5,000 & 716 & 1,009 & 2,287 & 822 & 666 \\ 
          & tokens & 11,477 & 75,216 & 17,585 & 92,004 & 17,002 & 19,913 & 28,783 & 8,336 & 10,675 \\ 
          \hline
        \end{tabular}
      }\end{center}
  }
  \caption{\small Statistics about the different datasets used in our
    experiments for the training (``train''), development (``dev'')
    and test (``test'') sets. ``sent.'' denotes the number of
    sentences in the dataset, ``tokens'' denotes the total number of
    words in the dataset, ``lex. size'' denotes the vocabulary size in
    the training set and ``\#nts'' denotes the number of nonterminals
    in the training set after binarization. \label{table:data}}
\end{table*}

The search algorithm can be used with any training algorithm for
L-PCFGs, including the algorithms of \newcite{cohen-13b} and
\newcite{narayan-15}. These methods, in their default setting, use a
function $f_{\s}$ which maps each nonterminal $a$ to a fixed number of
latent states $m_a$ it uses. In this case, $\s$ takes as input
training data, in the form of a treebank, decomposes into inside and
outside trees at each node in each tree in the training set; and
reduces the dimensionality of the inside and outside feature functions
by running SVD on the cross-covariance matrix $\Omega_a$ of the inside
and the outside trees, for each nonterminal $a$. \newcite{cohen-13b}
estimate the parameters of the L-PCFG up to a linear transformation
using $f(a)$ non-zero singular values of $\Omega_a$, whereas
\newcite{narayan-15} use the feature representations induced from the
SVD step to cluster instances of nonterminal $a$ in the training data
into $f(a)$ clusters; these clusters are then treated as latent states
that are ``observed.'' Finally, Narayan and Cohen follow up with a
simple frequency count maximum likelihood estimate to estimate the
parameters in the L-PCFG with these latent states.

An important point to make is that the learning algorithms of
\newcite{narayan-15} and \newcite{cohen-13b} are relatively
fast,\footnote{It has been documented in several papers that the
  family of spectral estimation algorithms is faster than algorithms
  such as EM, not just for L-PCFGs. See, for example,
  \newcite{parikh2012spectral}.} in comparison to the EM algorithm.
They require only one iteration over the data. In addition, the SVD
step of $\s$ for these learning algorithms is computed just once for a
large $m$. The SVD of a lower rank can then be easily computed from
that SVD.


\section{Experiments}
\label{sec:exp}

In this section, we describe our setup for parsing experiments on a
range of languages.

\subsection{Experimental Setup}

\paragraph{Datasets}
We experiment with nine treebanks consisting of eight different
morphologically rich languages: Basque, French, German, Hebrew,
Hungarian, Korean, Polish and Swedish. Table~\ref{table:data} shows
the statistics of 9 different treebanks with their splits into
training, development and test sets. Eight out of the nine datasets
(Basque, French, German-T, Hebrew, Hungarian, Korean, Polish and
Swedish) are taken from the workshop on Statistical Parsing of
Morphologically Rich Languages (SPMRL; Seddah et al.,
2013)\nocite{seddah-etal:2013:spmrl}. The German corpus in the SPMRL
workshop is taken from the TiGer corpus (German-T, Brants et al.,
2004\nocite{tiger2004}). We also experiment with another German
corpus, the NEGRA corpus (German-N, Skut et al., 1997\nocite{skut97}),
in a standard evaluation split.\footnote{We use the first 18,602
  sentences as a training set, the next 1,000 sentences as a
  development set and the last 1,000 sentences as a test set. This
  corresponds to an 80\%-10\%-10\% split of the treebank.} Words in
the SPMRL datasets are annotated with their morphological signatures,
whereas the NEGRA corpus does not contain any morphological
information.


\paragraph{Data preprocessing and treatment of rare words}
We convert all trees in the treebanks to a binary form, train and run
the parser in that form, and then transform back the trees when doing
evaluation using the PARSEVAL metric. In addition, we collapse unary
rules into unary chains, so that our trees are fully binarized. The
column ``\#nts'' in Table~\ref{table:data} shows the number of
nonterminals after binarization in the various treebanks. Before
binarization, we also drop all functional information from the
nonterminals. We use fine tags for all languages except Korean. This
is in line with \newcite{bjorkelund-etal:2013:spmrl}.\footnote{In
  their experiments \newcite{bjorkelund-etal:2013:spmrl} found that
  fine tags were not useful for Basque also; they did not find a
  proper explanation for that. In our experiments, however, we found
  that fine tags were useful for Basque.  To retrieve the fine tags,
  we concatenate coarse tags with their refinement feature (``AZP'')
  values.} For Korean, there are 2,825 binarized nonterminals making
it impractical to use our optimization algorithm, so we use the coarse
tags.




\newcite{bjorkelund-etal:2013:spmrl} have shown that the morphological
signatures for rare words are useful to improve the performance of
the Berkeley parser. In our preliminary experiments with na\"{i}ve
spectral estimation, we preprocess rare words in the training set in
two ways: (i) we replace them with their corresponding POS tags, and (ii)
we replace them with their corresponding POS+morphological
signatures. 
We follow \newcite{bjorkelund-etal:2013:spmrl} and consider a word to
be rare if it occurs less than 20 times in the training data. We
experimented both with a version of the parser that does not ignore and does ignore
letter cases, and discovered that the parser
behaves better when case is not ignored.

\begin{table*}[ht]
  {
    \begin{center}{\small
        \begin{tabular}{|l|l||c|c|c|c|c|c|c|c|c|}
          \hline 
          
          \multicolumn{2}{|c||}{lang.} & Basque & French & German-N & German-T & Hebrew &
          Hungarian & Korean & Polish & Swedish \\ \hline
          \parbox[t]{1mm}{\multirow{2}{*}{\rotatebox[origin=c]{90}{Bk}}}
          & van & 69.2 & \textbf{79.9} & - & 81.7 & 87.8 & 83.9 & 71.0 & 84.1 & 74.5 \\
          & rep & \textbf{84.3} & 79.7 & - & 82.7 & 89.6 & \textbf{89.1} & \textbf{82.8} & 87.1 & \textbf{75.5} \\ \hline
          \parbox[t]{1mm}{\multirow{3}{*}{\rotatebox[origin=c]{90}{Cl}}} 
          & van (pos) & 69.8 & {73.9} & {75.7} & 78.3 & 88.0 & 81.3 & 68.7 & 90.3 & 70.9 \\ 
          & van (rep) & {78.6} & 73.7 & - & {78.8} & {88.1} & {84.7} & {76.5} & {90.4} & {71.4} \\
          & opt & 81.2$^{*}$ & 76.7 & 77.8 & 81.7 & 90.1 & 87.2 & 79.2 & \textbf{92.0}$^{*}$ & 75.2 \\ \hline
          \parbox[t]{1mm}{\multirow{2}{*}{\rotatebox[origin=c]{90}{Sp}}} 
          & van & 78.1 & 78.0 & 77.6 & 82.0 & 89.2 & 87.7 & 80.6 & 91.7 & 73.4 \\
          & opt & 79.0 & 78.1$^{*}$ & \textbf{79.0}$^{*}$ & \textbf{82.9}$^{*}$ & \textbf{90.3}$^{*}$ & 87.8$^{*}$ & 80.9$^{*}$ & 91.7 & \textbf{75.5}$^{*}$ \\ \hline \hline
          \multicolumn{2}{|c||}{Bk multiple} & 87.4 & 82.5 & - & 85.0 & 90.5 & 91.1 & 84.6 & 88.4 & 79.5 \\ 
          \multicolumn{2}{|c||}{Cl multiple} & 83.4 & 79.9 & 82.7 & 85.1 & 90.6 & 89.0 & 80.8 &  92.5 & 78.3 \\ \hline
          \multicolumn{2}{|c||}{Hall et al. '14} & 83.7 & 79.4 & - & 83.3 & 88.1 & 87.4 & 81.9 & 91.1 & 76.0 \\
          \multicolumn{2}{|c||}{Crabb\'{e} '15} & 84.0 & 80.9 & - & 84.1 & 90.7 & 88.3 & 83.1 & 92.8 & 77.9 \\ \hline
        \end{tabular}
      }\end{center}
  }
  \caption{\small Results on the development datasets. ``Bk'' makes
    use of the Berkeley parser with its coarse-to-fine mechanism to
    optimize the number of latent states \cite{petrov-06}.  For Bk,
    ``van'' uses the vanilla treatment of rare words using signatures
    defined by \protect\newcite{petrov-06}, whereas ``rep.''  uses the
    morphological signatures instead. ``Cl'' uses the algorithm of
    \protect\newcite{narayan-15} and ``Sp'' uses the algorithm of
    \protect\newcite{cohen-13b}. In Cl, ``van (pos)'' and ``van
    (rep)'' are vanilla estimations (i.e., each nonterminal is mapped
    to fixed number of latent states) replacing rare words by POS or
    POS+morphological signatures, respectively. The best of these two
    models is used with our optimization algorithm in ``opt''. For Sp,
    ``van'' uses the best setting for unknown words as Cl.  Best
    result in each column from the first seven rows is in bold. In
    addition, our best performing models from rows 3-7 are marked with
    $^{*}$. ``Bk multiple'' shows the best results with the multiple
    models using product-of-grammars procedure
    \cite{petrov2010products} and discriminative reranking
    \cite{charniak-05}. ``Cl multiple'' gives the results with
    multiple models generated using the noise induction and decoded
    using the hierarchical decoding \cite{narayan-15}. Bk results are
    not available on the development dataset for German-N. For others,
    we report Bk results from \newcite{bjorkelund-etal:2013:spmrl}. We
    also include results from \newcite{hall2014less} and
    \newcite{crabbe:2015:EMNLP}.
\label{table:dev}}
\end{table*}

\begin{table*}[ht]
  {
    \begin{center}{\small
        \begin{tabular}{|l|l||c|c|c|c|c|c|c|c|c|}
          \hline 
          
          \multicolumn{2}{|c||}{lang.} & Basque & French & German-N & German-T & Hebrew &
          Hungarian & Korean & Polish & Swedish \\ \hline
          \multicolumn{2}{|c||}{Bk} & 74.7 & \textbf{80.4} & \textbf{80.1} & \textbf{78.3} & 87.0 & 85.2 & 78.6 & 86.8 & 80.6 \\ \hline
          \parbox[t]{1mm}{\multirow{2}{*}{\rotatebox[origin=c]{90}{Cl}}} 
          & van & 79.6 & 74.3 & 76.4 & 74.1 & 86.3 & 86.5 & 76.5 & 90.5 & 76.4 \\
          & opt & \textbf{81.4}$^{*}$ & 75.6 & 78.0 & 76.0 & 87.2 & 88.4 & 78.4 & 91.2$^{*}$ & 79.4 \\ \hline
          \parbox[t]{1mm}{\multirow{2}{*}{\rotatebox[origin=c]{90}{Sp}}} 
          & van & 79.9 & 78.7 & 78.4 & 78.0 & 87.8 & 89.1 & \textbf{80.3} & \textbf{91.8} & 78.4 \\
          & opt & 80.5 & 79.1$^{*}$ & 79.4$^{*}$ & 78.2$^{*}$ & \textbf{89.0}$^{*}$ & \textbf{89.2}$^{*}$ & 80.0$^{*}$ & \textbf{91.8} & \textbf{80.9}$^{*}$ \\ \hline \hline
          \multicolumn{2}{|c||}{Bk multiple} & 87.9 & 82.9 & 84.5 & 81.3 & 89.5 & 91.9 & 84.3 & 87.8 & 84.9 \\ 
          \multicolumn{2}{|c||}{Cl multiple} & 83.4 & 80.4 & 82.7 & 80.4 & 89.2 & 89.9 & 80.3 & 92.4 & 82.8 \\ \hline
          \multicolumn{2}{|c||}{Hall et al. '14} & 83.4 & 79.7 & - & 78.4 & 87.2 & 88.3 & 80.2 & 90.7 & 82.0 \\
          \multicolumn{2}{|c||}{F\&M '15} & 85.9 & 78.8 & - & 78.7 & 89.0 & 88.2 & 79.3 & 91.2 & 82.8 \\
          \multicolumn{2}{|c||}{Crabb\'{e} '15} & 84.9 & 80.8 & - & 79.3 & 89.7 & 90.1 & 82.7 & 92.7 & 83.2 \\ \hline
        \end{tabular}
      }\end{center}
  }
  \caption{\small Results on the test datasets. ``Bk'' denotes the
    best Berkeley parser result reported by the shared task organizers
    \cite{seddah-etal:2013:spmrl}. For the German-N data, Bk results
    are taken from \newcite{petrov2010products}. ``Cl van'' shows the
    performance of the best vanilla models from Table~\ref{table:dev}
    on the test set. ``Cl opt'' and ``Sp opt'' give the result of our
    algorithm on the test set. We also include results from
    \newcite{hall2014less}, \newcite{crabbe:2015:EMNLP} and
    \newcite{fernandezgonzalez-martins:2015}.\label{table:test}}
\end{table*}

\paragraph{Spectral algorithms: subroutine choices} 
The latent state optimization algorithm will work with either the
clustering estimation algorithm of \newcite{narayan-15} or the
spectral algorithm of \newcite{cohen-13b}. In our setup, we first run
the latent state optimization algorithm with the clustering
algorithm. We then run the spectral algorithm once with the optimized
$f$ from the clustering algorithm. We do that because the clustering
algorithm is significantly faster to iteratively parse the development
set, because it leads to sparse estimates.

Our optimization algorithm is sensitive to the initialization of the
number of latent states assigned to each nonterminals as it
sequentially goes through the list of nonterminals and chooses latent
state numbers for each nonterminal, keeping latent state numbers for
other nonterminals fixed. In our setup, we start our search algorithm
with the best model from the clustering algorithm, controlling for all
hyperparameters; we tune $f$, the function which maps each nonterminal
to a fixed number of latent states $m$, by running the vanilla version
with different values of $m$ for different languages. Based on our
preliminary experiments, we set $m$ to 4 for Basque, Hebrew, Polish
and Swedish; 8 for German-N; 16 for German-T, Hungarian and Korean;
and 24 for French.



We use the same features for the spectral methods as in
\newcite{narayan-15} for German-N. For the SPMRL datasets we do
not use the head features. These require linguistic understanding of
the datasets (because they require head rules for propagating leaf
nodes in the tree), and we discovered that simple heuristics for
constructing these rules did not yield an increase in
performance.

We use the \texttt{kmeans} function in Matlab to do the clustering
for the spectral algorithm of \newcite{narayan-15}. We experimented
with several versions of $k$-means, and discovered that the version
that works best in a set of preliminary experiments is hard
$k$-means.\footnote{To be more precise, we use the Matlab function
  \texttt{kmeans} while passing it the parameter
  \texttt{`start'=`sample'} to randomly sample the initial centroid
  positions. In our experiments, we found that default initialization
  of centroids differs in Matlab14 (random) and in Matlab15
  (kmeans++). Our estimation performs better with random
  initialization.}

\paragraph{Decoding and evaluation}
For efficiency, we use a base PCFG without latent states to prune
marginals which receive a value less than $0.00005$ in the dynamic
programming chart. This is just a bare-bones PCFG that is estimated
using maximum likelihood estimation (with frequency count). The parser
takes part-of-speech tagged sentences as input.  We tag the German-N
data using the Turbo Tagger \cite{martins-10}. For the languages in
the SPMRL data we use the MarMot tagger of
\newcite{mueller-schmid-schutze:2013:EMNLP} to jointly predict the POS
and morphological tags.\footnote{See
  \newcite{bjorkelund-etal:2013:spmrl} for the performance of the
  MarMot tagger on the SPMRL datasets.} The parser itself can assign
different part-of-speech tags to words to avoid parse failure. This is
also particularly important for constituency parsing with
morphologically rich languages. It helps mitigate the problem of the
taggers to assign correct tags when long-distance dependencies are
present.

For all results, we report the $F_1$ measure of the PARSEVAL metric
\cite{black-91}. We use the \texttt{EVALB} program\footnote{We use
  \texttt{EVALB} from \url{http://nlp.cs.nyu.edu/evalb/} for the
  German-N data and its modified version from
  \url{http://dokufarm.phil.hhu.de/spmrl2014/doku.php?id=shared_task_description}
  for the SPMRL datasets.} with the parameter file
\texttt{COLLINS.prm} \cite{collins-99b} for the German-N data and the
SPMRL parameter file, \texttt{spmrl.prm}, for the SPMRL data
\cite{seddah-etal:2013:spmrl}.

In this setup, the latent state optimization algorithm terminates in
few hours for all datasets except French and German-T. The German-T
data has 762 nonterminals to tune over a large development set
consisting of 5,000 sentences, whereas, the French data has a high
average sentence length of 31.43 in the development set.\footnote{To
  speed up tuning on the French data, we drop sentences with length
  $>$46 from the development set, dropping its size from 12,35 to
  1,006.} 

Following \newcite{narayan-15}, we further improve our results by
using multiple spectral models where noise is added to the underlying
features in the training set before the estimation of each
model.\footnote{We only use the algorithm of \newcite{narayan-15} for
  the noisy model estimation. They have shown that decoding with noisy
  models performs better with their sparse estimates than the dense
  estimates of \newcite{cohen-13b}.}  Using the optimized $f$, we
estimate $80$ models for each of noise induction mechanisms in Narayan
and Cohen: Dropout, Gaussian (additive) and Gaussian
(multiplicative). To decode with multiple noisy models, we train the
MaxEnt reranker of \newcite{charniak-05}.\footnote{Implementation:
  \url{https://github.com/BLLIP/bllip-parser}.  More specifically, we
  used the programs {\tt extract-spfeatures}, {\tt cvlm-lbfgs} and
  {\tt best-indices}. {\tt extract-spfeatures} uses head features,
  we bypass this for the SPMRL datasets by creating a dummy {\tt
    heads.cc} file.  {\tt cvlm-lbfgs} was used with the default
  hyperparameters from the Makefile.\label{footnote:bllip}}
Hierarchical decoding with ``maximal tree coverage'' over MaxEnt models,
further improves our accuracy. See \newcite{narayan-15} for more
details on the estimation of a diverse set of models, and on decoding
with them.

\begin{table*}[ht]
  {
    \begin{center}{\small
        \begin{tabular}{|l||c|c|c|c|c|c|c|c|c|c|c|c|}
          \hline
          & \multicolumn{4}{c|}{preterminals} & \multicolumn{4}{c|}{interminals} & \multicolumn{4}{c|}{all} \\
          language & $\sum_i x_i$ & $\sum_i y_i$ & div. & \#nts  & $\sum_i x_i$ & $\sum_i y_i$ & div. & \#nts & $\sum_i x_i$ & $\sum_i y_i$ & div. & \#nts\\
          \hline
          Basque & 311 & 419 & 196 & 169 & 91 & 227 & 152 & 31 & 402 & 646 & 348 & 200 \\
          French & 839 & 715 & 476 & 108 & 1145 & 1279 & 906 & 114 & 1984 & 1994 & 1382 & 222 \\
          German-N & 425 & 567 & 416 & 109 & 323 & 578 & 361 & 99 & 748 & 1145 & 777 & 208 \\
          German-T & 1251 & 890 & 795 & 378 & 1037 & 1323 & 738 & 384 & \textbf{2288} & \textbf{2213} & 1533 & 762 \\
          Hebrew & 434 & 442 & 182 & 279 & 169 & 544 & 393 & 96 & 603 & 986 & 575 & 375 \\
          Hungarian & 457 & 415 & 282 & 87 & 186 & 261 & 129 & 25 & 643 & 676 & 411 & 112 \\
          Korean & 1077 & 980 & 547 & 331 & 218 & 220 & 150 & 21 & \textbf{1295} & \textbf{1200} & 697 & 352 \\ 
          Polish & 252 & 311 & 197 & 135 & 132 & 180 & 86 & 63 & 384 & 491 & 283 & 198 \\
          Swedish & 191 & 284 & 127 & 106 & 85 & 345 & 266 & 42 & 276 & 629 & 393 & 148 \\
          \hline
        \end{tabular}
      }\end{center}
  }
  \caption{\small A comparison of the number of latent states for the
    different nonterminals before and after running our latent state
    number optimization algorithm.  The index $i$ ranges over
    preterminals and interminals, with $x_i$ denoting the number of
    latent states for nonterminal $i$ with the vanilla version of the
    estimation algorithm and $y_i$ denoting the number of latent
    states for nonterminal $i$ after running the optimization
    algorithm. The divergence figure (``div.'') is a calculation of
    $\sum_i | x_i - y_i |$.\label{table:states-number} \vspace{-0.3cm}}
\end{table*}

\subsection{Results}
\label{sec:results}

Table~\ref{table:dev} and Table~\ref{table:test} give the results for
the various languages.\footnote{See more in
  \url{http://cohort.inf.ed.ac.uk/lpcfg/}.} Our main focus is on
comparing the coarse-to-fine Berkeley parser \cite{petrov-06} to our
method. However, for the sake of completeness, we also present results
for other parsers, such as parsers of \newcite{hall2014less},
\newcite{fernandezgonzalez-martins:2015} and
\newcite{crabbe:2015:EMNLP}.

In line with \newcite{bjorkelund-etal:2013:spmrl}, our preliminary
experiments with the treatment of rare words suggest that
morphological features are useful for all SPMRL languages except
French. Specifically, for Basque, Hungarian and Korean, improvements
are significantly large.

Our results show that the optimization of the number of latent states
with the clustering and spectral algorithms indeed improves these
algorithms performance, and these increases generalize to the test
sets as well. This was a point of concern, since the optimization
algorithm goes through many points in the hypothesis space of parsing
models, and identifies one that behaves optimally on the development
set -- and as such it could overfit to the development set. However,
this did not happen, and in some cases, the increase in accuracy of
the test set after running our optimization algorithm is actually
larger than the one for the development set.


While the vanilla estimation algorithms (without latent state
optimization) lag behind the Berkeley parser for many of the
languages, once the number of latent states is optimized, our parsing
models do better for Basque, Hebrew, Hungarian, Korean, Polish and
Swedish. For German-T we perform close to the Berkeley parser
(78.2 vs. 78.3). It is also interesting to compare the clustering
algorithm of \newcite{narayan-15} to the spectral algorithm of
\newcite{cohen-13b}. In the vanilla version, the spectral algorithm
does better in most cases. However, these differences are narrowed,
and in some cases, overcome, when the number of latent states is
optimized. Decoding with multiple models further improves our
accuracy. Our ``Cl multiple'' results lag behind ``Bk multiple.'' We believe this is
the result of the need of head features for the MaxEnt
models.\footnote{\newcite{bjorkelund-etal:2013:spmrl} also use the
  MaxEnt raranker with multiple models of the Berkeley parser, and in
  their case also the performance after the raranking step is not
  always significantly better. See footnote~\ref{footnote:bllip} on how we
  create dummy head-features for our MaxEnt models.}
%

Our results show that spectral learning is a viable alternative
to the use of expectation-maximization coarse-to-fine techniques. As we discuss
later, further improvements have been introduced to state-of-the-art
parsers that are orthogonal to the use of a specific estimation algorithm.
Some of them can be applied to our setup.



\subsection{Further Analysis}

In addition to the basic set of parsing results, we also wanted to
inspect the size of the parsing models when using the optimization
algorithm in comparison to the vanilla
models. Table~\ref{table:states-number} gives this analysis. In this
table, we see that in most cases, on average, the optimization
algorithm chooses to enlarge the number of latent states. However, for
German-T and Korean, for example, the optimization algorithm actually
chooses a smaller model than the original vanilla model.

\begin{table*}[ht]
  {
    \begin{center}{\scriptsize
        \begin{tabular}{|l|c|c|p{0.2cm}||l|c|c|p{0.2cm}||l|c|c|p{0.2cm}||l|c|c|p{0.3cm}|}
          \hline 
          preterminal & freq. & b. & a. & preterminal & freq. & b. &
          a. & preterminal & freq. & b. & a. & preterminal & freq. &
          b. & a.\\
          \hline
          PWAT & 64 & 2 & 2 & TRUNC & 614 & 8 & 1 & PIS & 1,628 & 8 & 8 & KON & 8,633 & 8 & 30 \\
          XY & 135 & 3 & 1 & VAPP & 363 & 6 & 4 & \textbf{\$*LRB*} & \textbf{13,681} & \textbf{8} & \textbf{6} & PPER & 4,979 & 8 & 100 \\
          NP$|$NN & 88 & 2 & 1 & PDS & 988 & 8 & 8 & ADJD & 6,419 & 8 & 60 & \textbf{\$.} & \textbf{17,699} & \textbf{8} & \textbf{3} \\
          VMINF & 177 & 3 & 5 & AVP$|$ADV & 211 & 4 & 11 & KOUS & 2,456 & 8 & 1 & APPRART & 6,217 & 8 & 15 \\
          PTKA & 162 & 3 & 1 & FM & 578 & 8 & 3 & PIAT & 1,061 & 8 & 8 & ADJA & 18,993 & 8 & 10 \\
          VP$|$VVINF & 409 & 6 & 2 & VVIMP & 76 & 2 & 1 & NP$|$PPER & 382 & 6 & 1 & APPR & 26,717 & 8 & 7 \\
          PRELAT & 94 & 2 & 1 & KOUI & 339 & 5 & 2 & VVPP & 5,005 & 8 & 20 & VVFIN & 13,444 & 8 & 3 \\
          AP$|$ADJD & 178 & 3 & 1 & VAINF & 1,024 & 8 & 1 & PP$|$PROAV & 174 & 3 & 1 & \textbf{\$,} & \textbf{16,631} & \textbf{8} & \textbf{1} \\
          APPO & 89 & 2 & 2 & PRELS & 2,120 & 8 & 40 & VAFIN & 8,814 & 8 & 1 & VVINF & 4,382 & 8 & 10 \\
          PWS & 361 & 6 & 1 & CARD & 6,826 & 8 & 8 & PTKNEG & 1,884 & 8 & 8 & ART & 35,003 & 8 & 10 \\
          KOKOM & 800 & 8 & 37 & NE & 17,489 & 8 & 6 & PTKZU & 1,586 & 8 & 1 & ADV & 15,566 & 8 & 8 \\
          VP$|$VVPP & 844 & 8 & 5 & PRF & 2,158 & 8 & 1 & VVIZU & 479 & 7 & 1 & PIDAT & 1,254 & 8 & 20 \\
          PWAV & 689 & 8 & 1 & PDAT & 1,129 & 8 & 1 & PPOSAT & 2,295 & 8 & 6 & NN & 68,056 & 8 & 12 \\
          APZR & 134 & 3 & 2 & PROAV & 1,479 & 8 & 10 & PTKVZ & 1,864 & 8 & 3 & VMFIN & 3,177 & 8 & 1 \\
          \hline
        \end{tabular}
      }\end{center}
  }
  \caption{\small A comparison of the number of latent states for each
    preterminal for the German-N model, before (``b.'') running
    the latent state number optimization algorithm and after running
    it (``a.''). Note that some of the preterminals denote unary rules
    that were collapsed (the nonterminals in the chain are separated
    by $|$). We do not show rare preterminals with b. and a. both being 1.
    \label{table:divergence} \vspace{-0.3cm}}
\end{table*}

We further inspected the behavior of the optimization algorithm for
the preterminals in German-N, for which the optimal model chose (on
average) a larger number of latent
states. Table~\ref{table:divergence} describes this analysis. We see
that in most cases, the optimization algorithm chose to decrease the
number of latent states for the various preterminals, but in some
cases significantly increases the number of latent
states.\footnote{Interestingly, most of the punctuation symbols, such as
  \texttt{\$$^*$LRB$^*$}, \texttt{\$.} and \texttt{\$,}, drop their
  latent state number to a significantly lower value indicating that
  their interactions with other nonterminals in the tree are minimal.}


Our experiments dispel another ``common wisdom'' about spectral
learning and training data size. It has been believed that spectral
learning do not behave very well when small amounts of data are
available (when compared to maximum likelihood estimation algorithms
such as EM) -- however we see that our results do better than the
Berkeley parser for several languages with small training datasets,
such as Basque, Hebrew, Polish and Hungarian. The source of this
common wisdom is that ML estimators tend to be
statistically ``efficient:'' they extract more information from the
data than spectral learning algorithms do. Indeed, there is no reason
to believe that spectral algorithms are statistically efficient.
However, it is not clear that indeed for L-PCFGs with the EM
algorithm, the ML estimator is statistically efficient either. MLE is
statistically efficient under specific assumptions which are not
clearly satisfied with L-PCFG estimation. In addition, when the model
is ``incorrect,'' (i.e. when the data is not sampled from L-PCFG, as
we would expect from natural language treebank data), spectral
algorithms could yield better results because they can mimic a higher
order model. This can be understood through HMMs. When estimating an
HMM of a low order with data which was generated from a higher order
model, EM does quite poorly. However, if the number of latent states
(and feature functions) is properly controlled with spectral
algorithms, a spectral algorithm would learn a ``product'' HMM, where
the states in the lower order model are the product of states of a
higher order.\footnote{For example, a trigram HMM can be reduced to a
  bigram HMM where the states are products of the original trigram
  HMM.}



State-of-the-art parsers for the SPMRL datasets improve the Berkeley
parser in ways which are orthogonal to the use of the basic estimation
algorithm and the method for optimizing the number of latent
states. They include transformations of the treebanks such as with
unary rules \cite{bjorkelund-etal:2013:spmrl}, a more careful handling
of unknown words and better use of morphological information such as
decorating preterminals with such information
\cite{bjorkelund-etal:2014:spmrl-sancl,szanto-farkas:2014:EACL}, with
careful feature specifications \cite{hall2014less} and
head-annotations \cite{crabbe:2015:EMNLP}, and other techniques.  Some
of these techniques can be applied to our case.

\section{Conclusion}
\label{sec:conclusion}

We demonstrated that a careful selection of the number of latent
states in a latent-variable PCFG with spectral estimation has a
significant effect on the parsing accuracy of the L-PCFG.  We
described a search procedure to do this kind of optimization, and
described parsing results for eight languages (with nine
datasets). Our results demonstrate that when comparing the
expectation-maximization with coarse-to-fine techniques to our
spectral algorithm with latent state optimization, spectral learning
performs better on six of the datasets.  Our results are comparable to
other state-of-the-art results for these languages.  Using a diverse
set of models to parse these datasets further improves the results.

\section*{Acknowledgments}

The authors would like to thank David McClosky for his help with
running the BLLIP parser and his comments on the paper and also the
three anonymous reviewers for their helpful comments. We also thank Eugene Charniak, DK
Choe and Geoff Gordon for useful discussions. Finally, thanks to
Djam\'{e} Seddah for providing us with the SPMRL datasets and to
Thomas M\"{u}ller and Anders Bj\"{o}rkelund for providing us the
MarMot models.  This research was supported by an EPSRC grant
(EP/L02411X/1) and an EU H2020 grant (688139/H2020-ICT-2015; SUMMA).

\bibliography{nlp}
\bibliographystyle{acl2016}

\end{document}